\pdfoutput=1

\documentclass[11pt]{article}

\usepackage[preprint]{acl}

\usepackage{times}
\usepackage{latexsym}
\usepackage{float}
\usepackage{fvextra}

\usepackage[T1]{fontenc}

\usepackage[utf8]{inputenc}

\usepackage{microtype}

\usepackage{inconsolata}

\usepackage{graphicx}

\usepackage{booktabs}
\usepackage{multirow}
\usepackage{tabularx}

\title{Perceive, Reflect, and Plan: Designing LLM Agent for Goal-Directed City Navigation without Instructions}

\author{
 \textbf{Qingbin Zeng\textsuperscript{1}},
 \textbf{Qinglong Yang\textsuperscript{1}},
 \textbf{Shunan Dong\textsuperscript{1}},
 \textbf{Heming Du\textsuperscript{2}},
\\
 \textbf{Liang Zheng\textsuperscript{2}},
 \textbf{Fengli Xu\textsuperscript{1}},
 \textbf{Yong Li\textsuperscript{1}},
\\
\\
 \textsuperscript{1}Department of Electronic Engineering, Tsinghua University,Beijing, China\\
 \textsuperscript{2}School of Computing, Australian National University, Australia
\\
 \small{
   \textbf{Correspondence:} \href{Fengli Xu:fenglixu@tsinghua.edu.cn}{fenglixu@tsinghua.edu.cn}
 }
}

\begin{document}

\maketitle

\begin{abstract}
    This paper considers a scenario in city navigation: an AI agent is provided with language descriptions of the goal location with respect to some well-known landmarks; By only observing the scene around, including recognizing landmarks and road network connections, the agent has to make decisions to navigate to the goal location without instructions. This problem is very challenging, because it requires agent to establish self-position and acquire spatial representation of complex urban environment, where landmarks are often invisible. In the absence of navigation instructions, such abilities are vital for the agent to make high-quality decisions in long-range city navigation. With the emergent reasoning ability of large language models (LLMs), a tempting solution is to prompt LLMs to ``react'' on each observation and make decisions accordingly. However, this solution has very poor performance that the agent often repeatedly visits same locations and make short-sighted, inconsistent decisions. To address these issues, this paper introduces a novel agentic workflow featured by its abilities to \textit{perceive}, \textit{reflect} and \textit{plan}. Specifically, we find LLaVA-7B can be fine-tuned to \textit{perceive} the approximate direction and distance of landmarks with sufficient accuracy for city navigation. Moreover, \textit{reflection} is achieved through a memory mechanism, where past experiences are stored and can be retrieved with current perception for effective decision argumentation. \textit{Planning} uses reflection results to produce long-term plans, which can avoid short-sighted decisions in long-range navigation. We show the designed workflow significantly improves urban navigation ability of the LLM agent compared with the state-of-the-art baselines. The code is available at \url{https://anonymous.4open.science/r/PReP-13B5}.
\end{abstract}


\section{Introduction}

Navigation in complex and unknown urban environment is an important task for artificial intelligent agents. This paper studies goal-directed agent navigation in the city environment, where an agent is provided with visual street view perception and textual goal location description in terms of relation to some well-known landmarks, \emph{e.g.}, ``the destination is approximately 300 meters northeast from the Skyscraper A''. The agent should visually identify the landmarks from street view images, use them as anchors to infer the direction of and distance from the goal, and plan a series of actions to navigate to the goal. The task is challenging because it requires the agent to be aware of its own location and acquire spatial understanding of complex urban environment, where landmarks are sometimes invisible. 
In the absence of navigation instructions and maps, self-constructed cognitive map is vital for the agent to perform well in navigation tasks.

\begin{figure}[t]
    \centering
    \includegraphics[width=0.95\columnwidth]{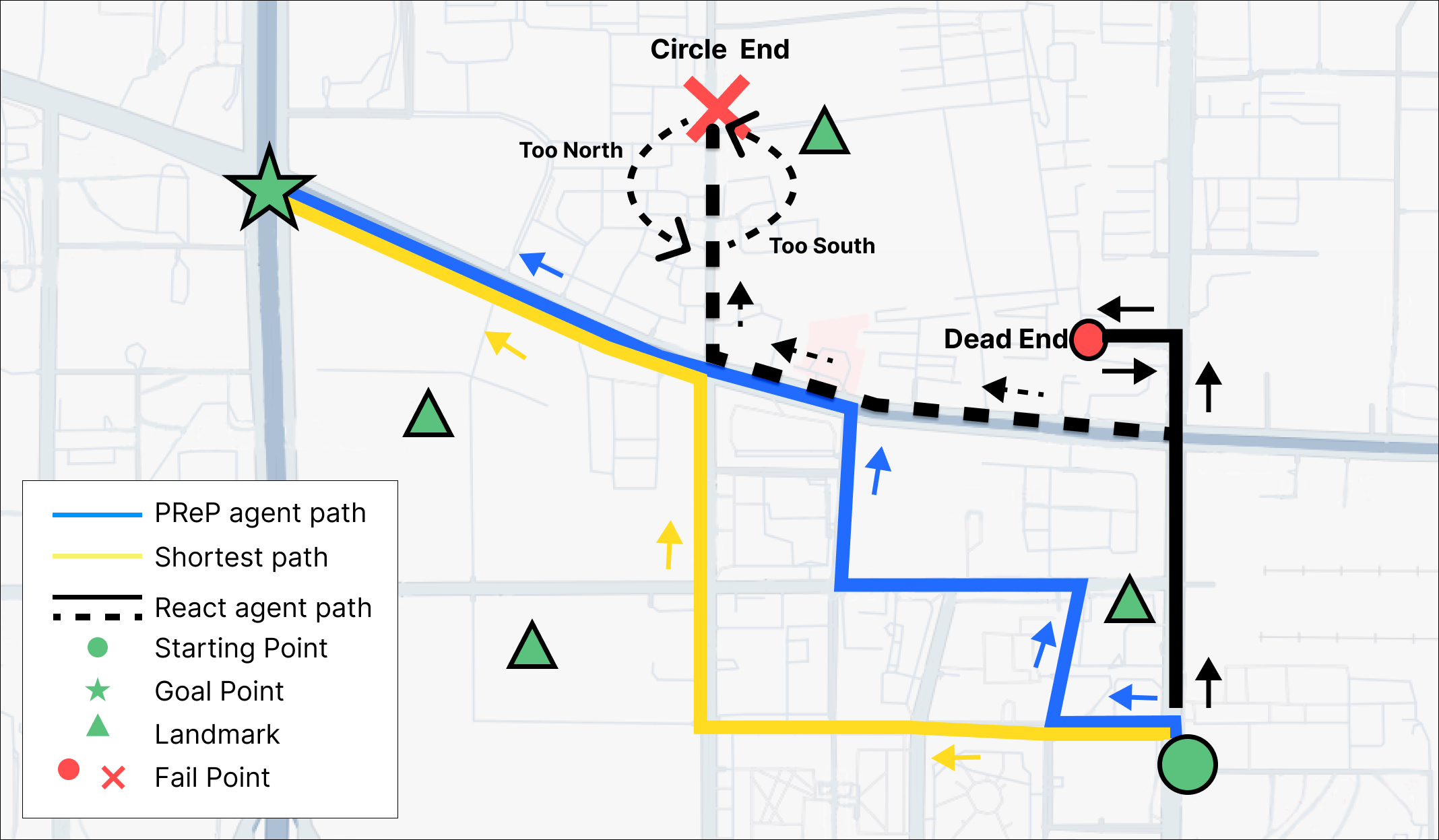} 
    \caption{\textbf{An illustrative comparison of city navigation results.} The proposed workflow method (blue) successfully reaches the goal, and its path is close to the shortest path (yellow). The React method (without workflow) fails because it makes short-sighted decisions. In one scenario, the React agent hits a dead end and keeps moving toward it because the goal is in that direction. In another scenario, the agent moves in circles because the goal's direction changes as it moves. The React agent have no memory so it can not take detours.}
    \label{fig:example}
\end{figure}

Existing literature does not provide a ready-to-use solution to this task. A few recent works~\cite{chen2019touchdown}~\cite{schumann2024velma} assume the availability of step-by-step language instructions and thus are not applicable to our task. Another branch of literature focus on designing reinforcement learning models~\cite{mirowski2016learning}~\cite{zhu2017target} \cite{wu2018building}, which often facing challenge of inefficient training samples and sensitivity to perturbations of the environment. 

We explore the use of large language models (LLMs) for this task. 
React~\cite{yao2022react} is a straightforward baseline to ground reasoning ability of LLMs in city environment. At each step, this method visually perceives the street views which is used to make an action decision. This process is iteratively performed until reaching the goal or running out of the navigation budget. While React has some success attempts in indoor environments, it performs poorly in complex urban environments, which can be attributed to two main reasons.   
First, because each action decision is based only on the current observation, the agent may repeat actions previously taken and find itself going around in circles. See Fig. \ref{fig:example} for an example. Second, React is short-sighted, focusing only on the immediate step. Without considering long-term action sequences, the agent would be prone to taking more actions than actually needed. To extend, the React without forming the cognitive map achieves such poor navigation performance in the complex environments.

This paper proposes an effective agentic workflow that improves the spatial cognitive ability of LLMs thus improving the goal-directed city navigation performance.
We fine-tune LLaVA and find it can perceive the direction and distance of landmarks with sufficient accuracy for navigation. 
Inspired by the theory of human cognition~\cite{sumers2023cognitive}, we propose a memory scheme to help the agent form the cognitive map~\cite{momennejad2023evaluatingcognitivemapsplanning}. The historical trajectories and observations are stored and summarized to learn an intrinsic spatial representation of the environment, \emph{i.e.}, an internal city map. The agent combines the historical experience and current observation to evaluate current situation and infer the goal direction. 
To improve over short-sighted actions, we resort to long-term planning. Specifically, considering the reflections and current road network connection, the agent decompose the full path into several sub-goals, ensuring consistent and reasonable movement to the final goal during long-range navigation. 
These components form the `\emph{Perceive}, \emph{Reflect}, and \emph{Plan}' workflow which allows the agent to perform long-range city navigation.

Our method only requires training the visual perception part with vision-language pairs of landmarks. The memory and planning parts both operate with few-shot examples (but also support fine-tuning). We find that fine-tuning the LLMs used in the memory and planning parts significantly improves navigation performance. For example, the fine-tuned LLaMA3 model achieves nearly a 13\% improvement in success rate compared to the base LLaMA3 model. Compared to RL methods, our approach offers a more data-efficient solution. Compared with instruction-following methods, our system does not rely on explicit instructions, allowing for greater autonomy in navigation. 

We collect navigation datasets reflecting CBD scenes in four cities——Beijing, Shanghai, New York and Paris. They contain complex road networks with thousands of road nodes and street view images. On the four datasets, the proposed workflow significantly outperforms methods that could be applied (but are not specific) to our task, achieving success rate of 54\% on the avergae of the four city test sets. We find the perception component produces accurate spatial relations to support city navigation, the success rate of which is only 5\% lower than navigation with ground truth perception results. 
Besides, we show that reflection and planning can help the LLM agent to form the cognitive map and further contribute to the success rate and make it useful when dealing with long-range navigation tasks.

\section{Related Work}
\textbf{Vision and language navigation} (VLN) aims to enable agents to autonomously navigate in visual environments based on natural language instructions~\cite{gu2022vision}\cite{wu2024vision}. The field evolves from indoor to urban settings, with expanded scope of tasks and datasets. Anderson \emph{et al.} created an early VLN dataset, while Mirowski \emph{et al.}~\cite{mirowski2018learning} introduce cross-modal matching models that leverage attention and reinforcement learning for vision and text integration. 
Recently, the use of LLMs has introduced new solutions in in VLN \cite{zhou2024navgpt}\cite{dorbala2023can}\cite{zu2023language}, which achieved success with indoor environment. Other works~\cite{shah2023lm}\cite{schumann2024velma} focusing on outdoor VLN,  which use strong language understanding capabilities of LLMs for navigation based on ground-level instructions. 
In comparison, our paper proposes a LLM agentic workflow for goal-directed city navigation, which requires less training cost than RL methods and does not rely on language instructions like previous LLM-based methods. 

\textbf{Agentic workflows with LLMs.} The exploration of agentic workflows using LLMs emerges as a effective strategy for planning problems~\cite{sumers2023cognitive}. Agentic workflows emphasize a step-by-step refinement process rather than single-step output generation. 
Studies such as inner monologue~\cite{huang2022inner} and reflexion~\cite{shinn2024reflexion} demonstrate the effectiveness of reflection in enhancing agentic understanding and reducing errors. Meanwhile, interactive planning methods like DEPS~\cite{wang2023describe} and RAP~\cite{hao2023reasoning} enable more structured and conscious planning. Other workflows, including CaP~\cite{liang2023code}, ProgPrompt~\cite{singh2023progprompt}, CoT~\cite{wei2022chain} and ToT~\cite{yao2024tree}, contribute to the collective understanding of how LLMs can be directed towards goal-oriented tasks. These methods are mostly used for mathematical and common reasoning problems, but whether LLMs can handle spatial reasoning, especially in navigation problems, hasn't been fully studied. In our work, we design an agentic workflow with perception, reflection, and planning modules, exploring the spatial cognitive abilities of LLMs to tackle complex urban navigation tasks.

\section{Task Definition and Dataset}
\label{section:env}

\begin{figure*}[ht]
    \centering
    \includegraphics[width=0.9\textwidth]{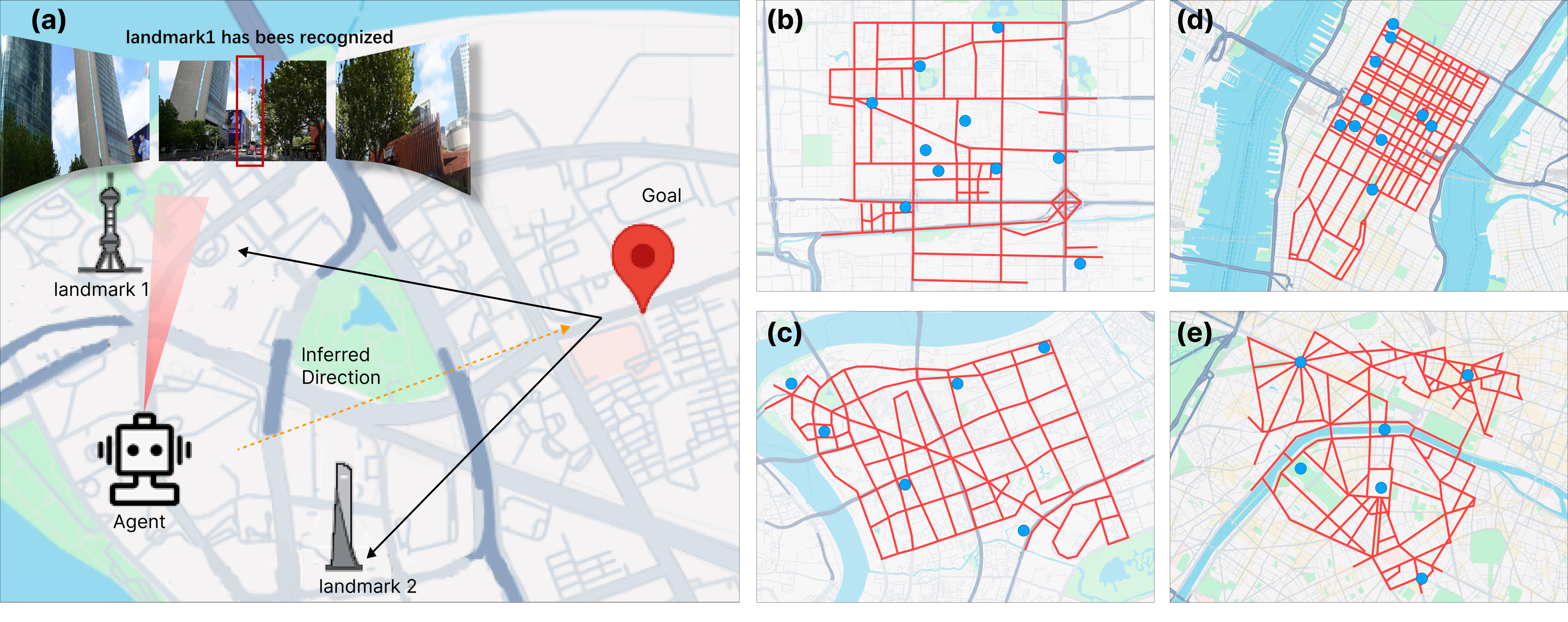} 
    \caption{\textbf{Task example and dataset regions.} A task example is shown in \textbf{(a)}. The instruction to the agent is the relative location of the goal \textit{w.r.t} the landmarks in the city environment. The agent perceives the street views and recognize the landmarks. Then the agent has to infer the goal position relative to its current location using its observations of landmarks and move through the urban space.
    The road networks are from chosen CBD areas in Beijing \textbf{(b)}, Shanghai \textbf{(c)}, New York \textbf{(d)} and Paris \textbf{(e)}. Blue points represent the landmarks while red lines are roads. }
    \label{fig:task_dataset}
\end{figure*}

\subsection{Task Definition}
In this study, an agent navigates in the urban environment to find the goal with visual perception and textual goal description. To define the task exactly, we give following definitions:

\textit{Definition 1 (Urban Environment)} The urban environment for navigation task can be described as an undirected graph $G=<V,E>$. Each node $v_i\in V$ represents an position in the road, while street views $S_{i}=\{s_i^1, s_i^2,...,s_i^k\}_{v_i}$ are the attached visual information for node $v_i$. The edge $e_{ij}\in E$ donates the moving path between $v_i$ and $v_j$. Define $E_i$ as the collection of all edges connect to the node $v_i$. What's more, landmarks $LM=\{lm_1, lm_2, ..., lm_n\} \in V$ are defined as isolated vertex in the urban environment $G$. As the graph corresponds to a real urban scene, and the relative positions of each node are fixed. We define the relative position relationship of $v_i$ with respect to $v_j$ as $R(v_i,v_j)$.

\textit{Definition 2 (Urban Navigation Task)} The urban navigation task can be formulated as finding a path from the start node $v_s$ to the goal node $v_g$ in the graph $G$. To be exact, given a navigation task $T=<v_s, v_g, D>$, the target is to find a shortest path to navigate to the goal. Description $D=\{R_1,R_2\}$ is to determine the goal, containing $R_1=\{R(v_g,lm)|lm\in LM_s\}$ and $R_2=\{R(lm_i,lm_j)|lm_i,lm_j\in LM\}$, where $R_1$ is the relative position between goal and landmarks visible in goal nodes, and $R_2$ is the relative position among all landmarks in the environment.

\textit{Definition 3 (Agent for Urban Navigation Task)} At timestamp $t$ and node $v_t$, the agent makes a decision to move to the next node $v_{t+1} = agent(T,S_t, E_t)$, utilizing the street views and road connectivity. The agent recognize the landmarks from the street views and then infer the goal direction and distance $R(v_t, v_g)$. Utilizing the above information, the agent navigates in the city street to find a path to the goal $p=[v_s,...,v_t,v_{t+1},...,v_g]$.

By studying this task, we not only seek to improve the success rate of navigation, but also hope to explore the spatial cognitive abilities of LLM.

\subsection{Dataset}

We collect data from central business districts (CBDs) of 4 cities, with a radius of a few kilometers. From this range, road network data are extracted and discritized at intervals of 50 meters forming the urban environment $G$. 
Each node of the road network is associated with the corresponding street view images. The number of street view images is equal to the degree of the node. Several well-known buildings are chosen as the landmarks in each area. 
Road network visualization and a task example are shown in Fig. \ref{fig:task_dataset}. 
More details about the datasets are in Appendix \ref{appendix:datasets}.

\section{PReP Agentic Workflow}
\label{section: PReP}

\begin{figure*}[ht]
    \centering
    \includegraphics[width=0.8\textwidth]{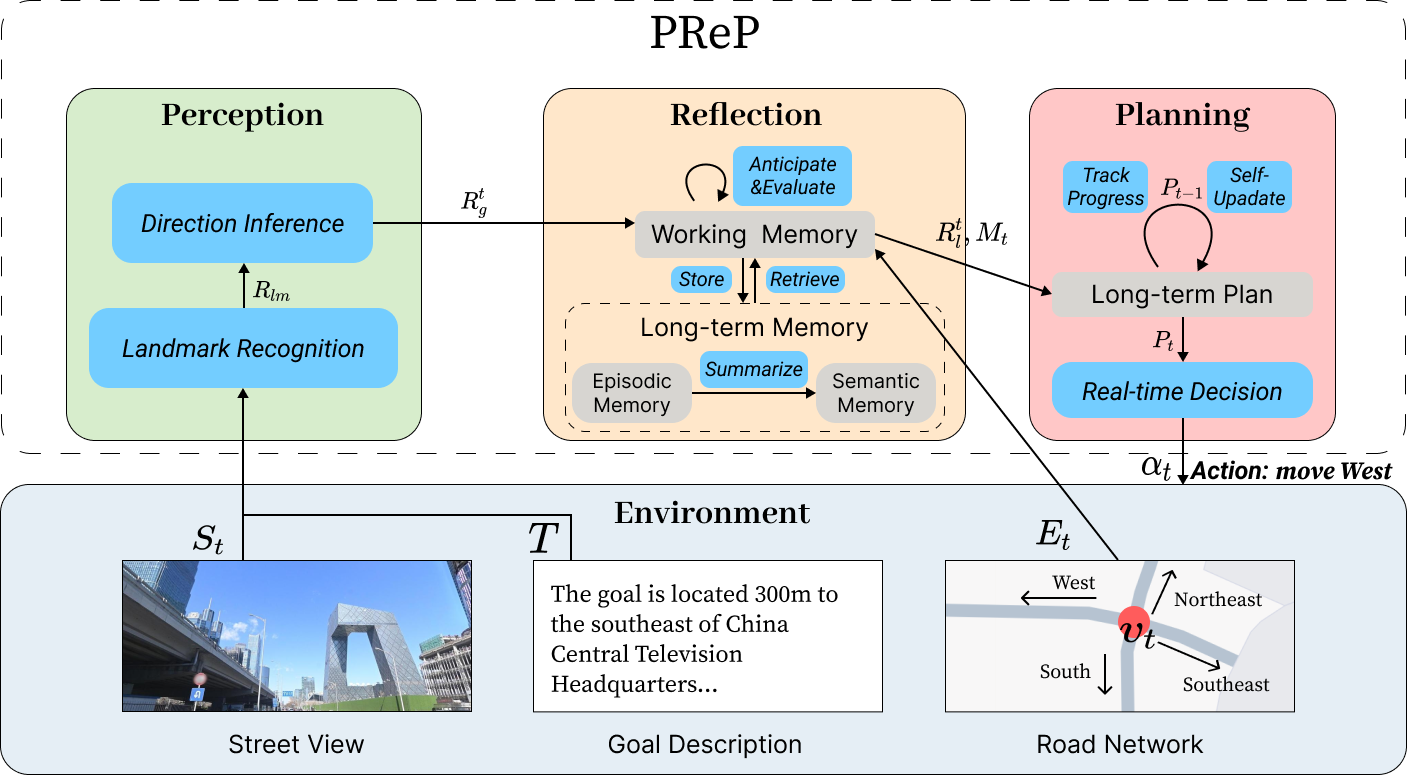} 
    \caption{\textbf{Overview of PReP workflow.} 
    It has three steps: perception, reflection, and planning. Blue boxes represent LLMs or LLaVA, while gray boxes indicate variables stored by natural language. Symbols are defined in section \ref{section:env}.
    }
    \label{fig:workflow}
\end{figure*}

\subsection{Workflow Overview}

The propoesd agentic workflow is shown in the Figure \ref{fig:workflow}, which is consist of three parts: visual perception, reflection with memory, and planning. While visual perception uses LLaVA, both reflection and planning uses large language models (LLMs). As described in Section \ref{subsection:perception}, \textit{visual perception} allows the agent to recognize the landmarks in the street view images and predict the direction and distance of the target. Perception results are passed to the \textit{reflection} part, where the agent reevaluates the perception results and reflects the goal location. In reflection, long-term memory is set up to summarize and learn from the historical trajectory for constructing intrinsic map representations, a topological map between nodes that the agent has visited. The \textit{planning} module is the decision core of the workflow. It generates a navigation plan by considering both the  goal direction after reflection and the current road connections. The agent follows this plan to make the next move. Then the agent state will be updated to start next iteration. These steps compose a workflow that enables the agent to form cognitive map of the urban environment and perform proper navigation actions. As shown in Section \ref{section: exp}, this workflow yields significantly higher success rates compared to the `React' baseline. 

\subsection{Perception} 
\label{subsection:perception}

The first part of the workflow is to equip the agent with visual perception ability. Although there are some neural network methods for processing the streetviews and recognizing the landmarks~\cite{vilera_landmark_2020}~\cite{boiarov_large_2019}, we choose to fine-tune the LLaVA-7B model for our perception component. There are two reasons for this: first, LLaVA is a multimodal LLM, making it easier to process both text and images at the same time; second, with the extensive world knowledge in LLMs, we can achieve better performance with fewer training samples.

\textbf{Perceiving the landmarks.} 
At timestamp $t$ and node $v_t$, the agent gets the street views $S_t=\{s_t^1,s_t^2,..s_t^k\}$. The agent then detects the landmarks and estimates their direction and distances $R_{lm}=\{R(lm_i, v_t)|lm_i\in LM\}$ to the agent itself. We perform the task using LLaVA \cite{liu2024visual}. Zero-shot LLaVA has poor recognition accuracy because the landmarks we use are probably not in its training data. We therefore fine-tunes LLaVA using the LoRA method \cite{hu2021lora}. To do so, we collect 5,000 landmark images and generate 30k Q\&A conversation data. More details can be seen in the Appendix \ref{appendix:llava}.

\textbf{Inferring directions to the goal.} 
While the agent acquires the landmarks position to the agent $R_{lm}$, it is necessary to infer the goal direction combining $R_{lm}$ and task description $T$. We can use an LLM to perform this inference. The output $R_g^t=R(v_g,v_t)$ of is the direction and distance of the goal relative to the agent. The essence of this problem is the cosine theorem, so the results can be further improved using calculators.

\subsection{Reflection}

Reflection is critical in our workflow, which summarizes past experience and reflects on visual perception results. 
This step has two main components: long-term memory and working memory. Long-term memory consists of episodic memory and semantic memory, where episodic memory stores navigation data and semantic memory saves summary of history navigation experience. Working memory serves as a data buffer to process the visual perception results and retrieved memory.
During the navigation process, the agent must have self-position and spatial understanding of the urban environment. It does not just rely on current perception but also uses historical trajectories. That's why we design the long-term memory component. Since current perception can sometimes be lacking and past movements are not always accurate, we design the anticipate-evaluate mechanism in working memory to make the agent more robust.

\textbf{Episodic memory.} 
Episodic memory is a list of the navigation data in natural language. When the agent moves from $v_{t}$ to $v_{t+1}$, this action and the perception results $R_g^t$ in $v_t$ are processed into a sentence and stored. The detail of the sentence format is in Appendix \ref{appendix:prompt}. Since the past navigation data are stored, the agent can retrieve the goal inference at history location and detect whether connected nodes are visited. For example, when the agent has saved navigation data in $v_t$ and navigated to $v_t^{'}$, it can retrieve $R_g^t$ and the status of connected nodes $E_{t^{'}}$ to help reflection and planning. 

\textbf{Semantic memory.} 
While episode memory records the experiences, the agent uses LLMs to summarize and learn from the episodic memory to form the semantic memory. The semantic memory is a high-level cognitive function that assists the agent in constructing an intrinsic representation of the navigation map. Like a human, it can understand the environment based on historical experience and learn more advanced navigation strategies, such as detours required to reach the destination. These strategies can be retrieved to working memory and further beneficial to the planning process. As the agent navigates, episodic memory and semantic memory are updated accordingly.

\textbf{Working memory.} 
Working memory receives visual perception results $R_g^t$ and retrieves relevant experiences from long-term memory. It has an anticipation-reevaluation mechanism to solve the problem where the agent cannot detect any landmarks in the street views and loses goal direction. Specifically, the agent uses historical perception results  $R_g^{t^{'}}$ and moving direction to anticipate the potential goal direction $R_p^{t}$, if any landmark can be observed, the agent reevaluates whether the current inference $R_g^t$ is reasonable, and synthesizes a new inference $R_l^t = LLM_{reflect}(R_g^t, R_p^t)$. The output of working memory to the planning part are the synthesized inference $R_l^t$ and retrieved memory. This enables the agent to tackle complex environments regardless whether landmarks are visible or not, making the agent more flexible and robust.

\begin{figure*}[ht]
  \centering
  \includegraphics[width=0.85\textwidth]{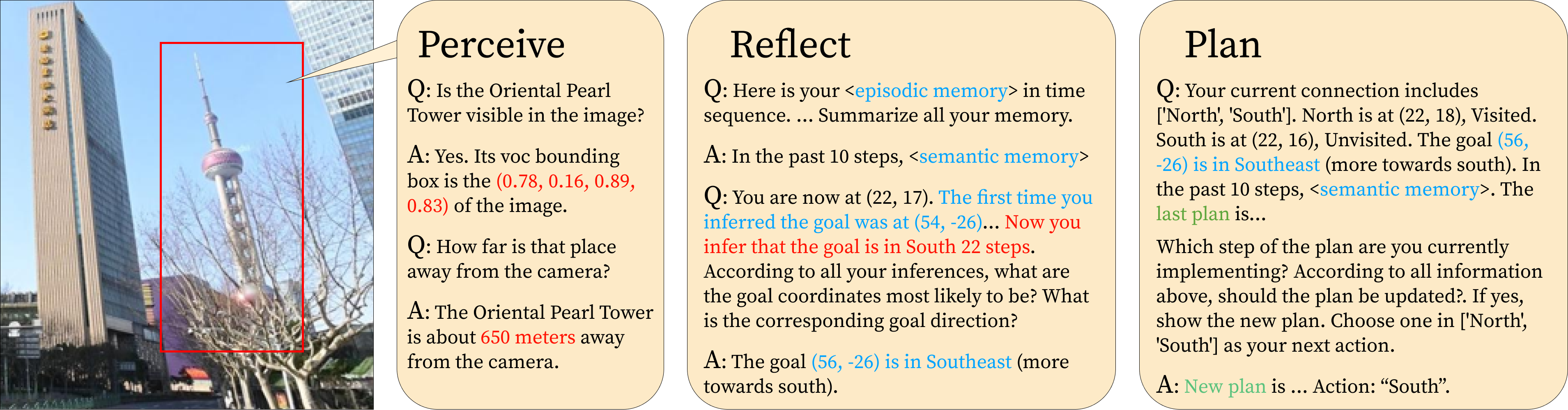}
  \caption{\textbf{Sample prompts and responses in the PReP workflow.} 
   In perception, a vision language model locates the landmarks and estimates their distances to the agent.  
  In reflection, the agent reflects on past memory, and gives an estimate of the direction of the goal.
  In planning, the agent uses the output from reflection to update the plan. Prompts have been simplified while retaining their original meaning. 
  The full prompts are provided in Appendix \ref{appendix:prompt}.
  }
  \label{fig:prompt}
\end{figure*}

\subsection{Planning}
Instead of reacting directly to observations, we use a planning module in our workflow. It involves long-term planning and short-term decision-making. Specifically, long-term planning uses reflected goal inference $R_l^t$, retrieved memory $M_t$, and the old plan $P_{t-1}$ as input. It updates the navigation plan at timestamp $t$ $P_t = LLM_{plan}(P_{t-1}, R_l^t, M_t)$ by analysing the old plan and breaking down the possible path into sub-goals. The agent firstly analyses what stage does the old plan has been carried out, and then synthesizes the goal inference, retrieved memory and the connection status to decide whether the plan should be updated. If so, the agent updates the plan by predicting the possible route to the goal, and breaking down the full route into few sub-goals like \textit{``move [east] until [an intersection]''}. 
After that, the short-term decision maker translates the plan into action $\alpha_t = LLM_{action}(P_t, E_t)$ based on the road connections $E_t$. The action $\alpha_t$ means a move from node $v_t$ to node $v_{t+1}$, making the agent updates its location and explore the goal in the environment.

\section{Experiments}
\label{section: exp}
\subsection{Experimental setup}

We experimentally evaluated the performance of the proposed agentic workflow on the simulated urban navigation task described in section \ref{section:env}. We use success rate (SR) and success rate weighted by path length (SPL) to measure system effectiveness and efficiency, respectively \cite{anderson2018evaluation}. 

All experiments were conducted on the same four test sets in the four cities. Each test set have 100 different navigation tasks with different start-goal pairs. 
Each start-goal pair is randomly chosen, and sometimes it is in a location where no landmarks are visible, making the task challenging.
The minimum number of steps required from the starting node to the goal follows a normal distribution with $\mu$=30 steps and $\sigma$=10 steps. Because one step translates to 50 meters on the map, it means the average navigation distance is 1,500 meters. We set the iteration limit as 2.5 times as the minimum steps. If the agent moves more than the limited steps, it is considered a failed task.

\begin{table*}[htbp]
  \small
  \centering
    \begin{tabularx}{\textwidth}{
        >{\hsize=1.5\hsize\centering\arraybackslash}X
        >{\hsize=0.5\hsize\centering\arraybackslash}X
        >{\hsize=0.5\hsize\centering\arraybackslash}X
        >{\hsize=0.5\hsize\centering\arraybackslash}X
        >{\hsize=0.5\hsize\centering\arraybackslash}X
        >{\hsize=0.5\hsize\centering\arraybackslash}X
        >{\hsize=0.5\hsize\centering\arraybackslash}X
        >{\hsize=0.5\hsize\centering\arraybackslash}X
        >{\hsize=0.5\hsize\centering\arraybackslash}X}
    \toprule
    &
      \multicolumn{2}{c}{Beijing} &
      \multicolumn{2}{c}{Shanghai} &
      \multicolumn{2}{c}{New York} &
      \multicolumn{2}{c}{Paris}
      \\
\cmidrule{2-9}  Methods & SR(\%) &  SPL(\%)  & SR(\%) &  SPL(\%)  &  SR(\%) &   SPL(\%) & SR(\%) &   SPL(\%)   \\
    \midrule
        Random  &  1 &  0.49 &  1 &  0.68 &   0 &     0 &  2 &  1.38   \\
        RL\cite{mirowski2018learning}   & 13 &  9.42 & 13 &  9.15 &   8 &  6.63 & 10 &  7.58   \\
    \midrule
        CoT     & 43 & 31.47 & 28 & 19.68 &   9 &  6.68 & 11 &  7.29   \\
        IM      & 40 & 31.46 & 26 & 21.27 &  19 & 15.27 & 15 &  12.17  \\
    Progprompt  & 44 & 32.75 & 24 & 18.45 &  10 &  7.45 & 21 &  16.02  \\
    Cap         & 47 & 35.86 & 23 & 17.13 &  12 &  9.11 & 17 &  12.31  \\
    DEPS        & 45 & 33.07 & 28 & 20.52 &  17 & 13.87 & 24 &  18.95  \\

    React       & 41 & 31.28 & 25 & 17.32 &   8 &  6.07 & 11 &  7.82   \\
        \midrule
    w/o Reflection      & 47 & 35.02 & 27 & 20.08 &  20 & 14.45 & 18 &  13.49  \\
    w/o Planning       & 56 & 42.71 & 46 & 33.66 &  41 & 31.07 & 42 &  32.00  \\
   \midrule
    PReP (ours) & \textbf{66} & \textbf{47.54} & \textbf{51} & \textbf{39.36} & \textbf{47} & \textbf{32.9} & \textbf{49} & \textbf{33.56} \\
    \bottomrule
    \end{tabularx}%
    \caption{Main results. We compare both LLM-free and LLM-based methods and perform an ablation study to evaluate the proposed reflection and planning methods.}
  \label{tab:main_results}%
\end{table*}%


\subsection{Main Evaluation}

\textbf{Comparison with existing methods that might give a solution.} 
We compared PReP with existing language-based methods, including Code as Policies (CaP)~\cite{liang2023code}, ProgPrompt~\cite{singh2023progprompt}, Inner Monologue (IM)~\cite{huang2022inner}, Chain of Thought (CoT)~\cite{wei2022chain}, DEPS~\cite{wang2023describe}, and React~\cite{yao2022react}. These methods use well-designed prompts adapted to our urban navigation task, and we provide the prompts for each method in Appendix \ref{appendix:prompt}. We also implemented two non-LLM baselines. The `random' method selects a random direction from the current connection each time. Reinforcement learning (RL) method is modified from ~\citet{mirowski2018learning} and trained for 1 million steps in the environment to learn the policy for reaching the goal. The perception module is the same for all methods. All the language-based methods use GPT-4-turbo as the base model, and all the hyper-parameters of LLMs are the same for a fair comparison. 


From Table \ref{tab:main_results}, we clearly observe that PReP yields the best navigation performance compared with existing methods across all cities. We have two key observations. \textbf{First}, the success rate of Random is nearly 0, indicating the significant challenge of this task. Existing language-based methods have improved performance, suggesting that LLMs possess the capability to navigate in cities based on goal direction. 
\textbf{Second}, in Beijing, PReP achieves an SR of 66.68\% and an SPL of 48.25\%, significantly surpassing competing methods, including DEPS (45\% SR, 33.07\% SPL) and Cap (47\% SR, 35.86\% SPL). This pattern holds in the other three cities, confirming PReP's effectiveness.

\textbf{Effectiveness of the proposed planning and reflection methods.}
We conduct ablation studies to validate the effectiveness of the reflection and planning methods. The perception component remains unchanged for all variants. Results are shown in Table \ref{tab:main_results}. `PReP' represents the full PReP workflow, which includes both planning and reflection. `w/o Reflection' means the agent follows a plan to find the goal, but does not reflect on memory or past inferences. `w/o Planning' means the agent receives reflected goal inference and retrieves memory but makes decisions without forming a long-term plan. `React' can also refers to PReP without planning or reflection, where the agent makes decisions based only on perception.

\begin{table*}[htbp]
\small
  \centering
    \begin{tabularx}{\textwidth}{
        >{\hsize=1.5\hsize\centering\arraybackslash}X
        >{\hsize=0.5\hsize\centering\arraybackslash}X
        >{\hsize=0.5\hsize\centering\arraybackslash}X
        >{\hsize=0.5\hsize\centering\arraybackslash}X
        >{\hsize=0.5\hsize\centering\arraybackslash}X
        >{\hsize=0.5\hsize\centering\arraybackslash}X
        >{\hsize=0.5\hsize\centering\arraybackslash}X
        >{\hsize=0.5\hsize\centering\arraybackslash}X
        >{\hsize=0.5\hsize\centering\arraybackslash}X}
    \toprule
       & \multicolumn{2}{c}{Beijing} & \multicolumn{2}{c}{Shanghai} & \multicolumn{2}{c}{New York} & \multicolumn{2}{c}{Paris} \\
\cmidrule{2-9}  Methods     & SR(\%) & SPL(\%) & SR(\%) & SPL(\%) & SR(\%) & SPL(\%) & SR(\%) & SPL(\%) \\
    \midrule
    LLaVA & 18 & 13.11 & 27 & 17.43 & 18 & 11.24 & 16 & 11.19 \\
    LLaVA-FT(Ours) & 66 & 47.54 & 51 & 39.36 & 47 & 32.9 & 49 & 33.56 \\
    \midrule
    Oracle & 70 & 53.17 & 54 & 40.41 & 52 & 34.03 & 53 & 34.71 \\
    \bottomrule
    \end{tabularx}%
  \caption{Comparisons among variants of the perception part}
  \label{tab:perception}%
\end{table*}%

\begin{table*}[htbp]
\small
  \centering
    \begin{tabularx}{\textwidth}{
        >{\hsize=1.5\hsize\centering\arraybackslash}X
        >{\hsize=0.5\hsize\centering\arraybackslash}X
        >{\hsize=0.5\hsize\centering\arraybackslash}X
        >{\hsize=0.5\hsize\centering\arraybackslash}X
        >{\hsize=0.5\hsize\centering\arraybackslash}X
        >{\hsize=0.5\hsize\centering\arraybackslash}X
        >{\hsize=0.5\hsize\centering\arraybackslash}X
        >{\hsize=0.5\hsize\centering\arraybackslash}X
        >{\hsize=0.5\hsize\centering\arraybackslash}X}
    \toprule
       & \multicolumn{2}{c}{Beijing} & \multicolumn{2}{c}{Shanghai} & \multicolumn{2}{c}{New York} & \multicolumn{2}{c}{Paris} \\
\cmidrule{2-9}    LLMs   & SR(\%) & SPL(\%) & SR(\%) & SPL(\%) & SR(\%) & SPL(\%) & SR(\%) & SPL(\%) \\
    \midrule
    GPT3.5-turbo & 33 & 21.62 & 17 & 10.87 & 10 & 6.04 & 14 & 9.08 \\
    GLM\_4 & 36 & 23.78 & 21 & 18.16 & 20 & 17.75 & 22 & 18.41 \\
    Mistral-7B & 8  & 4.22 & 3  & 2.15 & 2  & 0.86 & 9  & 4.7 \\
    LLaMA3-8B & 32 & 19.65 & 15 & 10.63 & 16 & 11.51 & 18 & 12.73 \\
    \midrule
    LLaMA3-FT & 45 & 31.3 & 36 & 22.38 & 23 & 17.95 & 27 & 18.33 \\
    \midrule
    GPT4-turbo & \textbf{66} & \textbf{47.54} & \textbf{51} & \textbf{39.36} & \textbf{47} & \textbf{32.9} & \textbf{49} & \textbf{33.56} \\
    \bottomrule
    \end{tabularx}%
  \caption{Performance of PReP using different LLMs}
  \label{tab:LLM}%
\end{table*}%

The full system clearly performs best. For example, on the Beijing test set, it achieves a success rate 25\%, 19\%, and 10\% higher than `React', `w/o Reflection', and `w/o Planning', respectively. This indicates the necessity of having both steps in our system. We also observe a larger performance drop when reflection is removed compared to when planning is removed, suggesting that reflection plays a critical role in cognitive map formation and navigation performance improvement.

\begin{figure*}[!htb]
    \centering
    \includegraphics[width=0.9\textwidth]{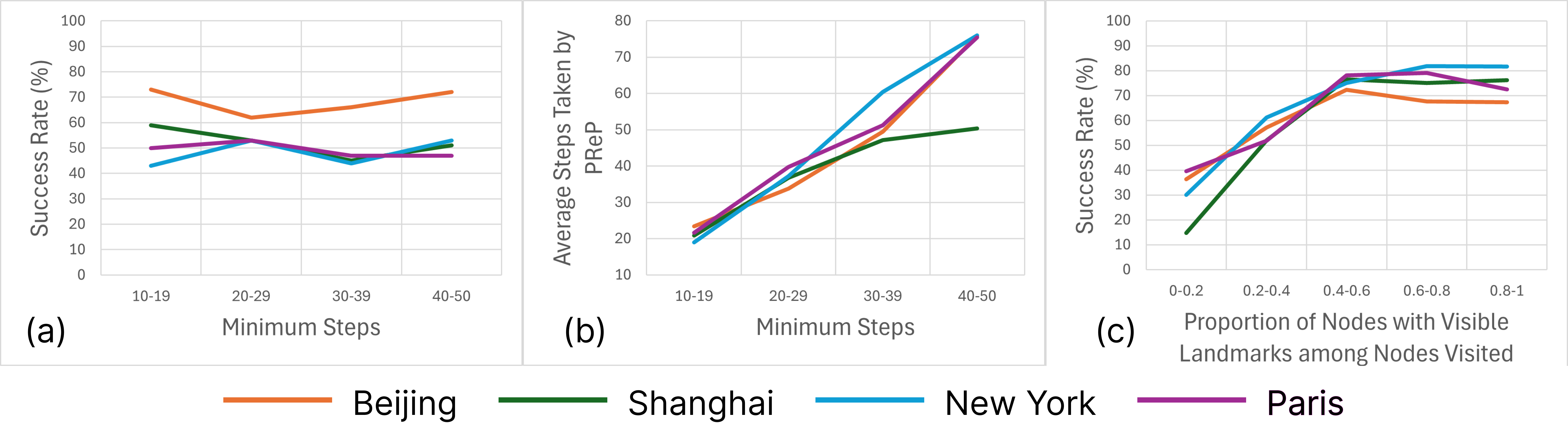}
    \caption{Performance of PReP across varying task difficulties
    }
    \label{fig6}
\end{figure*}

\subsection{Further Analysis}
\label{section: fur_exp}
\textbf{Benefit of fine-tuning LLaVA over zero-shot LLaVA.} In Table \ref{tab:perception}, we compare fine-tuned LLaVA with zero-shot LLaVA. Zero-shot LLaVA has much poorer performance: on the Shanghai test sets, its SR and SPL is 27\% and 23\% lower than its fine-tuned version. It indicates that LLaVA does not naturally recognize landmarks through street views. But interestingly, zero-shot LLaVA still has 10+\% success rate. This can be explained by its 19\% accuracy, 6\% precision, 93\% recall, and 0.64\$ IoU (See Appendix \ref{appendix:llava}). In fact, zero-shot LLaVA has good building detection capacity and assume that most images contain landmarks, leading to a high recall. Its precision is low (6\%), but sometimes is fine for the agent to find the goal. 
  
We also compare with an oracle setting, where the perception results are replaced with ground truth directions and distances measured by GPS. Compared with oracle results, fine-tuned LLaVA is average 4\% lower in SR and average 2.5\% lower in SPL, respectively. This is not a significant gap, indicating the effectiveness of fine-tuning. 

\textbf{Comparing different LLMs.} We use different LLMs to perform inference (blue boxes in Fig.\ref{fig:prompt}). These models include GPT-3.5-turbo, GLM-4~\cite{zhipu2024glm4}, Mistral-7B~\cite{jiang2023mistral}, LLaMA3-8B~\cite{meta2024llama3} and GPT-4-turbo. From Table \ref{tab:LLM}, we observe that GPT-4-turbo significantly outperforms other LLMs without fine-tuning. Moreover, we then use the question-answering data generated by GPT-4-turbo to fine-tune LLaMA3~\cite{zheng2024llamafactory} (see Appendix \ref{appendix:llama}). The fine-tuned LLaMA3 achieves performance that was second only to GPT-4-turbo among all models.

\textbf{Impact of goal distance.} 
We analyze whether the distance between the goal and the starting point has an impact on success rate. In Fig. \ref{fig6}a, we observe that the SR does not decrease a lot when the goal is as far as 2 kilometers (40 - 50 steps) from the starting point. Instead, the SR remains relatively stable across various distances. The possible reason is that the maximum number of iteration limit increases as the distance increases, the agent may fully explore the environment and find the goal more easily. It may also suggest that the task difficulty is not significantly correlated with distance, but rather with the complexity of the road network.
Furthermore, in Fig. \ref{fig6}b, we study how the goal distance affects the number of steps taken by PReP. Basically the agent takes steps that is about 1.5 times the minimum steps to reach the goal.

\textbf{Impact of landmark visibility.} 
When the agent performs a navigation task, some nodes it passes through can observe landmarks, while others cannot. We study how landmark visibility impacts its success rate. In Fig. \ref{fig6}c, it can be observed that the success rate increases significantly with the increase in the proportion of nodes along the path where landmarks are visible. When more than 40\% of the nodes on the path can observe landmarks, the success rate gradually converges, which might suggest that such landmark visibility is adequate to support the agent in accomplishing the navigation task. It is worth noting that in the Beijing test set, the agent could identify landmarks in approximately 50\% of all nodes. In other test sets, this percentage may drop to less than 20\%. This may have led to a significant higher success rate on the Beijing test set compared to other test sets in Fig. \ref{fig6}a, indicating that successfully identifying landmarks from the environment is a key factor for the agent to navigate to the goal.

\section{Conclusion}
We propose an agentic workflow for goal-directed city navigation without step-by-step language instructions or maps. The workflow includes a fine-tuned LLaVA model for spatial perception, a memory module for synthesizing and reflecting perception results and retrieved memory, and a planning module for navigation route planning. As our approach only requires training the visual perception part, it is a more data-efficient solution compared to RL methods. Owing to the well-designed reflection and planning part, the agent can perform the long-term navigation task in complex environment and achieves a success rate of about 54\%. Further experiments show that the agent performs well in 4 cities and various difficulty levels, demonstrating robustness and flexibility. Our contributions not only present an effective agentic workflow for using LLMs in goal-directed urban navigation, but also validate the potential of LLMs for complex spatial tasks.

\section{Limitations}
\label{section:limit}
Despite the advancements and innovative approaches in our work, there are several limitations to consider. A notable challenge is the dependence on powerful closed-source models, such as GPT-4-turbo, for superior results. Although we fine-tuned the open-source LLaMA-8B model, its performance, while better than other LLMs, does not reach half that of GPT-4-turbo. This highlights a significant gap. Future work should explore more effective fine-tuning strategies for open-source models to enhance their capabilities, ensuring our progress in urban navigation can be widely adopted and further developed.
Another issue is the limited size of our test set, which may cause fluctuations in the success rate. A larger and more diverse test set could provide a more comprehensive evaluation of the model's performance and robustness.

\bibliography{custom}

\begin{thebibliography}{32}
\providecommand{\natexlab}[1]{#1}

\bibitem[{Anderson et~al.(2018)Anderson, Chang, Chaplot, Dosovitskiy, Gupta,
  Koltun, Kosecka, Malik, Mottaghi, Savva et~al.}]{anderson2018evaluation}
Peter Anderson, Angel Chang, Devendra~Singh Chaplot, Alexey Dosovitskiy,
  Saurabh Gupta, Vladlen Koltun, Jana Kosecka, Jitendra Malik, Roozbeh
  Mottaghi, Manolis Savva, et~al. 2018.
\newblock On evaluation of embodied navigation agents.
\newblock \emph{arXiv preprint arXiv:1807.06757}.

\bibitem[{Boiarov and Tyantov(2019)}]{boiarov_large_2019}
Andrei Boiarov and Eduard Tyantov. 2019.
\newblock \href {https://doi.org/10.1145/3357384.3357956} {Large scale landmark
  recognition via deep metric learning}.
\newblock In \emph{Proceedings of the 28th {ACM} International Conference on
  Information and Knowledge Management}, {CIKM} '19, pages 169--178.
  Association for Computing Machinery.

\bibitem[{Chen et~al.(2019)Chen, Suhr, Misra, Snavely, and
  Artzi}]{chen2019touchdown}
Howard Chen, Alane Suhr, Dipendra Misra, Noah Snavely, and Yoav Artzi. 2019.
\newblock Touchdown: Natural language navigation and spatial reasoning in
  visual street environments.
\newblock In \emph{Proceedings of the IEEE/CVF Conference on Computer Vision
  and Pattern Recognition}, pages 12538--12547.

\bibitem[{Dorbala et~al.(2023)Dorbala, Mullen~Jr, and Manocha}]{dorbala2023can}
Vishnu~Sashank Dorbala, James~F Mullen~Jr, and Dinesh Manocha. 2023.
\newblock Can an embodied agent find your “cat-shaped mug”? llm-based
  zero-shot object navigation.
\newblock \emph{IEEE Robotics and Automation Letters}.

\bibitem[{Gu et~al.(2022)Gu, Stefani, Wu, Thomason, and Wang}]{gu2022vision}
Jing Gu, Eliana Stefani, Qi~Wu, Jesse Thomason, and Xin~Eric Wang. 2022.
\newblock Vision-and-language navigation: A survey of tasks, methods, and
  future directions.
\newblock \emph{arXiv preprint arXiv:2203.12667}.

\bibitem[{Hao et~al.(2023)Hao, Gu, Ma, Hong, Wang, Wang, and
  Hu}]{hao2023reasoning}
Shibo Hao, Yi~Gu, Haodi Ma, Joshua~Jiahua Hong, Zhen Wang, Daisy~Zhe Wang, and
  Zhiting Hu. 2023.
\newblock \href {https://arxiv.org/abs/2305.14992} {Reasoning with language
  model is planning with world model}.
\newblock \emph{Preprint}, arXiv:2305.14992.

\bibitem[{Hu et~al.(2021)Hu, Shen, Wallis, Allen-Zhu, Li, Wang, Wang, and
  Chen}]{hu2021lora}
Edward~J Hu, Yelong Shen, Phillip Wallis, Zeyuan Allen-Zhu, Yuanzhi Li, Shean
  Wang, Lu~Wang, and Weizhu Chen. 2021.
\newblock Lora: Low-rank adaptation of large language models.
\newblock \emph{arXiv preprint arXiv:2106.09685}.

\bibitem[{Huang et~al.(2022)Huang, Xia, Xiao, Chan, Liang, Florence, Zeng,
  Tompson, Mordatch, Chebotar et~al.}]{huang2022inner}
Wenlong Huang, Fei Xia, Ted Xiao, Harris Chan, Jacky Liang, Pete Florence, Andy
  Zeng, Jonathan Tompson, Igor Mordatch, Yevgen Chebotar, et~al. 2022.
\newblock Inner monologue: Embodied reasoning through planning with language
  models.
\newblock \emph{arXiv preprint arXiv:2207.05608}.

\bibitem[{Jiang et~al.(2023)Jiang, Sablayrolles, Mensch, Bamford, Chaplot,
  Casas, Bressand, Lengyel, Lample, Saulnier et~al.}]{jiang2023mistral}
Albert~Q Jiang, Alexandre Sablayrolles, Arthur Mensch, Chris Bamford,
  Devendra~Singh Chaplot, Diego de~las Casas, Florian Bressand, Gianna Lengyel,
  Guillaume Lample, Lucile Saulnier, et~al. 2023.
\newblock Mistral 7b.
\newblock \emph{arXiv preprint arXiv:2310.06825}.

\bibitem[{Liang et~al.(2023)Liang, Huang, Xia, Xu, Hausman, Ichter, Florence,
  and Zeng}]{liang2023code}
Jacky Liang, Wenlong Huang, Fei Xia, Peng Xu, Karol Hausman, Brian Ichter, Pete
  Florence, and Andy Zeng. 2023.
\newblock Code as policies: Language model programs for embodied control.
\newblock In \emph{2023 IEEE International Conference on Robotics and
  Automation (ICRA)}, pages 9493--9500. IEEE.

\bibitem[{Liu et~al.(2024)Liu, Li, Wu, and Lee}]{liu2024visual}
Haotian Liu, Chunyuan Li, Qingyang Wu, and Yong~Jae Lee. 2024.
\newblock Visual instruction tuning.
\newblock \emph{Advances in neural information processing systems}, 36.

\bibitem[{Meta(2024)}]{meta2024llama3}
Meta. 2024.
\newblock Introducing meta llama 3: The most capable openly available llm to
  date.
\newblock \url{https://ai.meta.com/blog/meta-llama-3/}.

\bibitem[{Mirowski et~al.(2018)Mirowski, Grimes, Malinowski, Hermann, Anderson,
  Teplyashin, Simonyan, Zisserman, Hadsell et~al.}]{mirowski2018learning}
Piotr Mirowski, Matt Grimes, Mateusz Malinowski, Karl~Moritz Hermann, Keith
  Anderson, Denis Teplyashin, Karen Simonyan, Andrew Zisserman, Raia Hadsell,
  et~al. 2018.
\newblock Learning to navigate in cities without a map.
\newblock \emph{Advances in neural information processing systems}, 31.

\bibitem[{Mirowski et~al.(2016)Mirowski, Pascanu, Viola, Soyer, Ballard,
  Banino, Denil, Goroshin, Sifre, Kavukcuoglu et~al.}]{mirowski2016learning}
Piotr Mirowski, Razvan Pascanu, Fabio Viola, Hubert Soyer, Andrew~J Ballard,
  Andrea Banino, Misha Denil, Ross Goroshin, Laurent Sifre, Koray Kavukcuoglu,
  et~al. 2016.
\newblock Learning to navigate in complex environments.
\newblock \emph{arXiv preprint arXiv:1611.03673}.

\bibitem[{Momennejad et~al.(2023)Momennejad, Hasanbeig, Vieira, Sharma, Ness,
  Jojic, Palangi, and Larson}]{momennejad2023evaluatingcognitivemapsplanning}
Ida Momennejad, Hosein Hasanbeig, Felipe Vieira, Hiteshi Sharma, Robert~Osazuwa
  Ness, Nebojsa Jojic, Hamid Palangi, and Jonathan Larson. 2023.
\newblock \href {https://arxiv.org/abs/2309.15129} {Evaluating cognitive maps
  and planning in large language models with cogeval}.
\newblock \emph{Preprint}, arXiv:2309.15129.

\bibitem[{Schumann et~al.(2024)Schumann, Zhu, Feng, Fu, Riezler, and
  Wang}]{schumann2024velma}
Raphael Schumann, Wanrong Zhu, Weixi Feng, Tsu-Jui Fu, Stefan Riezler, and
  William~Yang Wang. 2024.
\newblock Velma: Verbalization embodiment of llm agents for vision and language
  navigation in street view.
\newblock In \emph{Proceedings of the AAAI Conference on Artificial
  Intelligence}, volume~38, pages 18924--18933.

\bibitem[{Shah et~al.(2023)Shah, Osi{\'n}ski, Levine et~al.}]{shah2023lm}
Dhruv Shah, B{\l}a{\.z}ej Osi{\'n}ski, Sergey Levine, et~al. 2023.
\newblock Lm-nav: Robotic navigation with large pre-trained models of language,
  vision, and action.
\newblock In \emph{Conference on robot learning}, pages 492--504. PMLR.

\bibitem[{Shinn et~al.(2024)Shinn, Cassano, Gopinath, Narasimhan, and
  Yao}]{shinn2024reflexion}
Noah Shinn, Federico Cassano, Ashwin Gopinath, Karthik Narasimhan, and Shunyu
  Yao. 2024.
\newblock Reflexion: Language agents with verbal reinforcement learning.
\newblock \emph{Advances in Neural Information Processing Systems}, 36.

\bibitem[{Singh et~al.(2023)Singh, Blukis, Mousavian, Goyal, Xu, Tremblay, Fox,
  Thomason, and Garg}]{singh2023progprompt}
Ishika Singh, Valts Blukis, Arsalan Mousavian, Ankit Goyal, Danfei Xu, Jonathan
  Tremblay, Dieter Fox, Jesse Thomason, and Animesh Garg. 2023.
\newblock Progprompt: Generating situated robot task plans using large language
  models.
\newblock In \emph{2023 IEEE International Conference on Robotics and
  Automation (ICRA)}, pages 11523--11530. IEEE.

\bibitem[{Sumers et~al.(2023)Sumers, Yao, Narasimhan, and
  Griffiths}]{sumers2023cognitive}
Theodore~R Sumers, Shunyu Yao, Karthik Narasimhan, and Thomas~L Griffiths.
  2023.
\newblock Cognitive architectures for language agents.
\newblock \emph{arXiv preprint arXiv:2309.02427}.

\bibitem[{Vilera et~al.(2020)Vilera, Rachmadi, and
  Yuniarno}]{vilera_landmark_2020}
Reza Vilera, Reza~Fuad Rachmadi, and Eko~Mulyanto Yuniarno. 2020.
\newblock \href {https://doi.org/10.1109/CENIM51130.2020.9298008} {Landmark
  segmentation and selective feature extraction in street-view image}.
\newblock In \emph{2020 International Conference on Computer Engineering,
  Network, and Intelligent Multimedia ({CENIM})}, pages 440--444.

\bibitem[{Wang et~al.(2023)Wang, Cai, Chen, Liu, Ma, and
  Liang}]{wang2023describe}
Zihao Wang, Shaofei Cai, Guanzhou Chen, Anji Liu, Xiaojian Ma, and Yitao Liang.
  2023.
\newblock Describe, explain, plan and select: Interactive planning with large
  language models enables open-world multi-task agents.
\newblock \emph{arXiv preprint arXiv:2302.01560}.

\bibitem[{Wei et~al.(2022)Wei, Wang, Schuurmans, Bosma, Xia, Chi, Le, Zhou
  et~al.}]{wei2022chain}
Jason Wei, Xuezhi Wang, Dale Schuurmans, Maarten Bosma, Fei Xia, Ed~Chi, Quoc~V
  Le, Denny Zhou, et~al. 2022.
\newblock Chain-of-thought prompting elicits reasoning in large language
  models.
\newblock \emph{Advances in neural information processing systems},
  35:24824--24837.

\bibitem[{Wu et~al.(2024)Wu, Chang, Li, Yin, and Hu}]{wu2024vision}
Wansen Wu, Tao Chang, Xinmeng Li, Quanjun Yin, and Yue Hu. 2024.
\newblock Vision-language navigation: a survey and taxonomy.
\newblock \emph{Neural Computing and Applications}, 36(7):3291--3316.

\bibitem[{Wu et~al.(2018)Wu, Wu, Gkioxari, and Tian}]{wu2018building}
Yi~Wu, Yuxin Wu, Georgia Gkioxari, and Yuandong Tian. 2018.
\newblock Building generalizable agents with a realistic and rich 3d
  environment.
\newblock \emph{arXiv preprint arXiv:1801.02209}.

\bibitem[{Yao et~al.(2024)Yao, Yu, Zhao, Shafran, Griffiths, Cao, and
  Narasimhan}]{yao2024tree}
Shunyu Yao, Dian Yu, Jeffrey Zhao, Izhak Shafran, Tom Griffiths, Yuan Cao, and
  Karthik Narasimhan. 2024.
\newblock Tree of thoughts: Deliberate problem solving with large language
  models.
\newblock \emph{Advances in Neural Information Processing Systems}, 36.

\bibitem[{Yao et~al.(2022)Yao, Zhao, Yu, Du, Shafran, Narasimhan, and
  Cao}]{yao2022react}
Shunyu Yao, Jeffrey Zhao, Dian Yu, Nan Du, Izhak Shafran, Karthik Narasimhan,
  and Yuan Cao. 2022.
\newblock React: Synergizing reasoning and acting in language models.
\newblock \emph{arXiv preprint arXiv:2210.03629}.

\bibitem[{Zheng et~al.(2024)Zheng, Zhang, Zhang, Ye, Luo, and
  Ma}]{zheng2024llamafactory}
Yaowei Zheng, Richong Zhang, Junhao Zhang, Yanhan Ye, Zheyan Luo, and Yongqiang
  Ma. 2024.
\newblock \href {http://arxiv.org/abs/2403.13372} {Llamafactory: Unified
  efficient fine-tuning of 100+ language models}.
\newblock \emph{arXiv preprint arXiv:2403.13372}.

\bibitem[{Zhipu(2024)}]{zhipu2024glm4}
Zhipu. 2024.
\newblock Zhipu ai devday glm-4.
\newblock \url{https://zhipuai.cn/en/devday}.

\bibitem[{Zhou et~al.(2024)Zhou, Hong, and Wu}]{zhou2024navgpt}
Gengze Zhou, Yicong Hong, and Qi~Wu. 2024.
\newblock Navgpt: Explicit reasoning in vision-and-language navigation with
  large language models.
\newblock In \emph{Proceedings of the AAAI Conference on Artificial
  Intelligence}, volume~38, pages 7641--7649.

\bibitem[{Zhu et~al.(2017)Zhu, Mottaghi, Kolve, Lim, Gupta, Fei-Fei, and
  Farhadi}]{zhu2017target}
Yuke Zhu, Roozbeh Mottaghi, Eric Kolve, Joseph~J Lim, Abhinav Gupta,
  Li~Fei-Fei, and Ali Farhadi. 2017.
\newblock Target-driven visual navigation in indoor scenes using deep
  reinforcement learning.
\newblock In \emph{2017 IEEE international conference on robotics and
  automation (ICRA)}, pages 3357--3364. IEEE.

\bibitem[{Zu et~al.(2023)Zu, Song, Chen, Guo, Sun, Tian, Pan, and
  Wang}]{zu2023language}
Weiqin Zu, Wenbin Song, Ruiqing Chen, Ze~Guo, Fanglei Sun, Zheng Tian, Wei Pan,
  and Jun Wang. 2023.
\newblock Language and sketching: An llm-driven interactive multimodal
  multitask robot navigation framework.
\newblock \emph{arXiv preprint arXiv:2311.08244}.

\end{thebibliography}

\appendix

\section{Ethical analysis}
\label{appendix:ethic}
The data we collected is open access including \emph{Baidu StreetViews API}, \emph{Google Streetviews API} and \emph{Open Street Map}, without privacy issues. Our ethical analysis confirms that all data was gathered in compliance with the code of ethics. We ensured that no personally identifiable information was collected, maintaining the anonymity and privacy of individuals. By adhering to the ethical guidelines, we avoid any potential privacy concerns and support the broader scientific community's efforts to build on and verify our work.

\section{Social impacts}
\label{appendix: impacts}
Developing the LLM agent for goal-directed city navigation has many positive social impacts. This technology extends beyond navigation robots, providing invaluable help to the visually impaired by enhancing their mobility and independence. Easier navigation in urban environments can greatly improve their quality of life and help them participate more fully in society. Additionally, using this technology in disaster relief can save lives by helping rescue teams navigate affected areas quickly and efficiently. However, there are potential negative social impacts to consider. Relying on AI for navigation might decrease human spatial awareness and problem-solving skills. To maximize the positive impact and minimize negative consequences, it is crucial to develop and implement this technology ethically with strong safeguards in place.

\section{Datasets details}
\label{appendix:datasets}
We collect data from central business districts (CBDs) of 4 cities, which have a radius of a few kilometers. From this range, road network data are extracted and discritized at intervals of 50 meters forming the urban environment. 
Each node of the road network is associated with the corresponding street view images. The number of street view images is equal to the degree of the node. Some features of the dataset are shown in Tab.\ref{tab:dataset}. 
Specifically, the selected area in Beijing is the Guomao CBD area, with a radius of approximately 3 kilometers, which includes a total of 1,134 nodes and 2,742 street view images, along with 10 landmark buildings. In Shanghai, the selected area is the Lujiazui CBD area, also with the similar radius, containing a total of 1,038 nodes and 2,366 street view images, along with 6 landmark buildings. While in New York area is with 1140 nodes and 2568 related street view images, Paris area have 1141 nodes and 2480 street view images, along with 10 and 6 landmarks respectively.
The landmarks chosen in the dataset are well-known buildings with unique features, such as the Oriental Pearl Tower in Shanghai and the Effiel Tower in Paris. Street view images are obtained from Baidu Map Street View API or Google Map API, with a field of view of 90° and an elevation angle of 20° for each image. The image resolution is 1,024×512 pixels.

\begin{table}[H]
\centering
\renewcommand{\arraystretch}{1.2} 
\begin{tabularx}{\columnwidth}{|X|c|c|c|}
\hline
City     & Nodes & Images &  Landmarks \\ \hline
Beijing  & 1134  & 2742   & 10         \\ \hline
Shanghai & 1038  & 2366   & 6          \\ \hline
New York & 1140  & 2568   & 10         \\ \hline
Paris    & 1141  & 2480   & 6          \\ \hline
\end{tabularx}
\caption{Dataset features.}
\label{tab:dataset}
\end{table}

\section{Additional Experimental Details}

\subsection{Fine-tuning LLaVA}
\label{appendix:llava}

The methodology for fine-tuning LLaVA uses the LoRA (Low-Rank Adaptation) technique. This approach introduces trainable low-rank matrices to simulate parameter updates, enabling rapid task adaptation without significantly increasing model complexity. We collected 250 images for each of the 32 landmarks from the Internet or the streetviews API, totaling 8,000 images. These images were manually annotated with the binary visibility and bounding box of each landmark. For those images collected from API, we can easiliy calculate the distance between the the landmarks and the carema using the metadata. We split the data into an 80\% training set and a 20\% test set. Using these images, we generated dialogue data in a question-and-answer format. The questions we asked are as following step by step, aiming to form a pattern of CoT to improve its understanding and reasoning ability.

\begin{enumerate}
    \item{"Is the $landmark_i$ visible in the image?"}
    \item{"The $landmark_i$ is visible in the image, what's the bounding box of it in the image?"}
    \item{"The $landmark_i$ is visible in the image and its bounding box is $(x_{min}, y_{min}, x_{max}, y_{max})$, how far is it actually away from the camera?"}
\end{enumerate}

We generated about 30k turn dialogues to fine-tune the llava-v1.5-7b. The fine-tuning process was carried out on one NVIDIA GeForce RTX A100 GPU and took about 3 hours. The scripts to fine-tuning is modified from the official repository \footnote{\url{https://github.com/haotian-liu/LLaVA}} with the default parameters.
 
\begin{figure}[!htbp]
    \centering
    \includegraphics[width=\columnwidth]{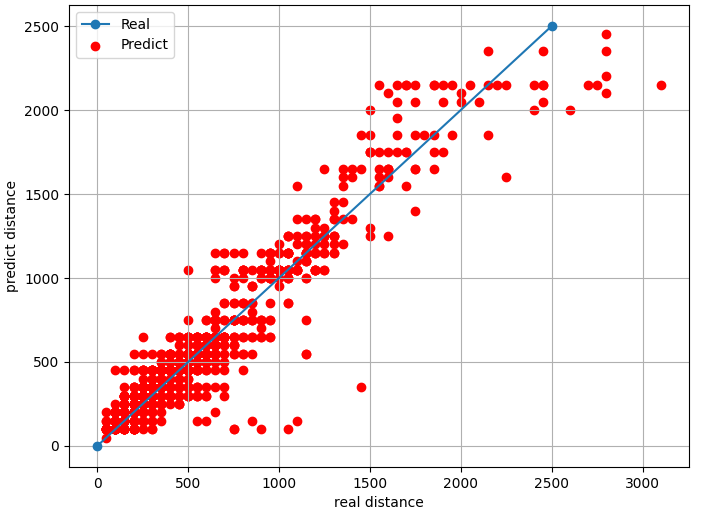}
    \caption{\textbf{The distance estimation results of fine-tuned LLaVA. }}
    \label{fig:llava_distance}
\end{figure}

The outcome of this fine-tuning process is a model that demonstrates remarkable accuracy in landmark recognition and segmentation (see Table \ref{tab:llava_ft}). While the distance estimation is not very accurate, the rough estimation results are still effective in subsequent navigation steps (see Fig \ref{fig:llava_distance}). The fine-tuned LLaVA model is essential for the agent to perceive the environment and obtain goal information.

\begin{table*}[htbp]
  \centering
  \begin{tabularx}{\textwidth}{>{\centering\arraybackslash}X >{\centering\arraybackslash}X >{\centering\arraybackslash}X >{\centering\arraybackslash}X >{\centering\arraybackslash}X >{\centering\arraybackslash}X}
    \toprule
     &
      Accuracy &
      Precision &
      Recall &
      F1-Score &
      IoU
      \\
    \midrule
    LLaVA-base &
      0.1873 &
      0.0576 &
      0.9347 &
      0.1072 &
      0.6432
      \\
    LLaVA-FT &
      \textbf{0.9980} &
      \textbf{0.9868} &
      \textbf{0.9695} &
      \textbf{0.9779} &
      \textbf{0.9152}
      \\
    \bottomrule
  \end{tabularx}%
  \caption{\textbf{The ability to recognize and segment landmark in the street view images.}}
  \label{tab:llava_ft}%
\end{table*}%

\subsection{Fine-tuning LLaMA3-8B}
\label{appendix:llama}

We are trying to transfer knowledge from a much large model (GPT-4-turbo) to a small model (LLaMA3-8B). The method involves using data generated during navigation with GPT-4-turbo to fine-tune LLaMA-8B. We filter the successful samples from all the saved data and process them into the ShareGPT format. To avoid data breaches, we use data generated in Beijing to fine-tune LLaMA, which is then tested in Shanghai, and vice versa. The situation of New York and Paris is similar. We generated about 20k dialogue turns for each city dataset and fine-tuned LLaMA-8B using the LoRA method with one NVIDIA GeForce RTX A100 GPU. This process took about 30 minutes using the LLaMA-Factory tool \footnote{\url{https://github.com/hiyouga/LLaMA-Factory/}}. All parameters in the fine-tuning process were set to default.

The results are shown in Table 4 in main content. We see that while LLaMA3-8B performs similarly to GPT-3.5-turbo in our task, it outperforms other LLMs (except GPT-4-turbo) after fine-tuning. Although there is still a gap compared to GPT-4-turbo, increasing the amount of fine-tuning data may improve its performance.

\section{Computational cost} 
The primary training cost is LLaVA fine-tuning, which requires one NVIDIA GeForce RTX A100 GPU with 80G memory for approximately 3 hours with 30k conversation data. Each request-response cycle of the fine-tuned LLaVA on the same GPU takes 6 to 8 seconds, while calling the LLM API takes 2 to 5 seconds (varies among different models). Each iteration for an agent step takes about 12 seconds. In future we will work to optimize the inference process. 

\begin{figure*}[!htbp]
    \centering
    \includegraphics[width=0.8\textwidth]{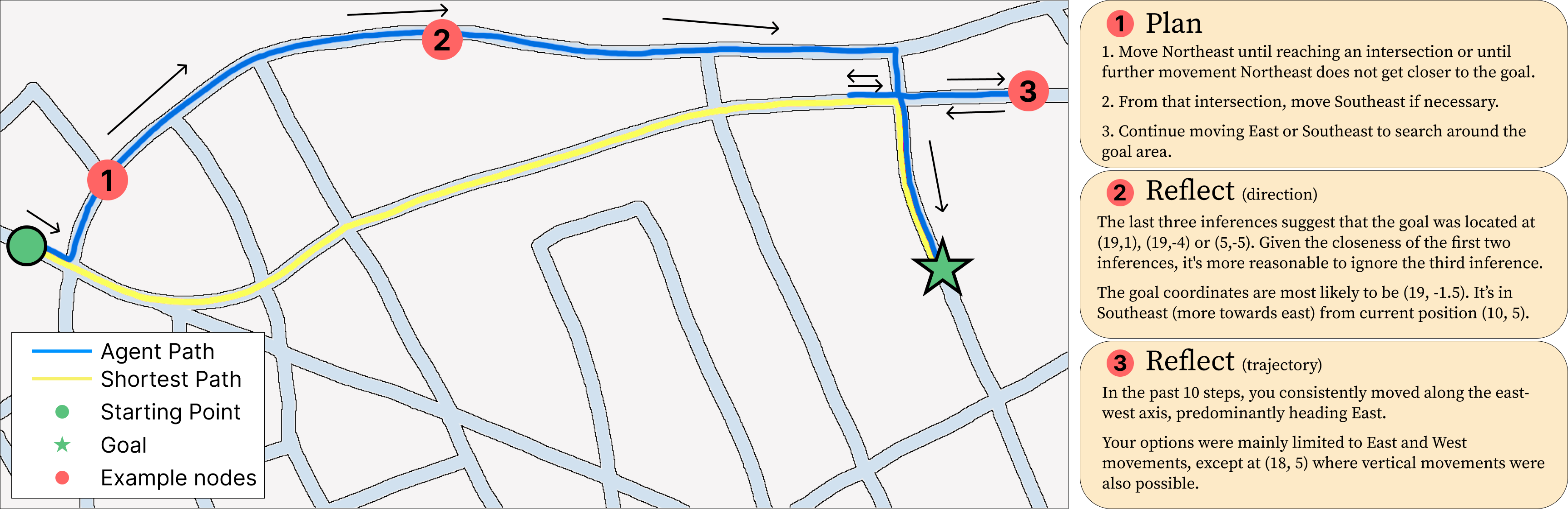}
    \caption{\textbf{Case study.} Here we choose three typical reasoning process in the whole navigation task where the arrows show the direction of travel of the agent. The agent makes a long-term plan at location (1) so that it can move consistently without moving back. At location (2), the agent cannot perceive landmarks in the street views and infer goal direction, but it can reflect history memory and anticipate the goal position. At location (3) the agent summarize its moving trajectory and find itself moving in the deviate direction, thus returning to the right route.}
  \label{fig:case_study}
\end{figure*}

\section{Case Study} 
We conduct a case study to illustrate the role of reflection and planning (see Fig. \ref{fig:case_study}). In this case, the PReP agent deviates from the shortest path in the beginning but still successfully reaches the goal. Initially, the agent infers that the goal was located to the east. However, as there was no direct path to due east, the agent planned a detour: it first heads northeast and then turns towards either east or southeast. The agent continuously makes inferences of the goal location each time it observes a landmark along its route. When the agent cannot perceive landmarks in the street views and lose the goal direction, it can reflect on history memory including the moving trajectory and goal inference, and then anticipate the goal direction from current position. When the agent moves in the deviate direction, it can reflect its trajectory and re-plan the right route.

\section{ Prompt example for different methods}
\label{appendix:prompt}

\subsection{Prompt of Full PReP}
\begin{Verbatim}[fontsize=\tiny, frame=single, samepage=false, breaklines=true, breakanywhere=true]
Perception
Q:
Is the Beijing_Zhoxin_Building visible in the image?

A:
Yes. Its voc bounding box is the (0.58984, 0.46680, 0.63672, 0.72461) of the image.

Q:
The Beijing_Zhoxin_Building is visible in the image and its voc bounding box is (0.58984, 0.46680, 0.63672, 0.72461). How far is that place away from the camera?

A:
The Beijing_Zhoxin_Building is about 1600 meters away from the camera.

Reflection
Q:Here is your memory list in time sequence.
1. You were at (14, 14). You could move to ['Northeast', 'North', 'West', 'South'] from there. You chose to move to Northeast. You then arrived at (15, 15).
2. You were at (15, 15). You could move to ['East', 'Northwest', 'Southwest'] from there. You chose to move to East. You then arrived at (16, 15).
3. You were at (16, 15). You could move to ['West', 'East'] from there. You chose to move to East. You then arrived at (17, 15).
4. You were at (17, 15). You could move to ['West', 'East'] from there. You chose to move to East. You then arrived at (18, 15).
5. You were at (18, 15). You could move to ['Northeast', 'West'] from there. You chose to move to Northeast. You then arrived at (19, 16).
6. You were at (19, 16). You could move to ['Northeast', 'Southwest'] from there. You chose to move to Northeast. You then arrived at (20, 17).
7. You were at (20, 17). You could move to ['Northeast', 'Southwest'] from there. You chose to move to Northeast. You then arrived at (21, 18).
8. You were at (21, 18). You could move to ['Northeast', 'Southwest'] from there. You chose to move to Northeast. You then arrived at (22, 19).
9. You were at (22, 19). You could move to ['South', 'East', 'Southwest', 'North'] from there. You chose to move to South. You then arrived at (22, 18).
10. You were at (22, 18). You could move to ['North', 'South'] from there. You chose to move to South. You then arrived at (22, 17).
Summarize all your memory, what can you learn from it?

A:
In the past 10 steps, you primarily moved in a northeastern direction, transitioning from (14, 14) to (22, 19), and then shifted to a southern direction, moving from (22, 19) to (22, 17). Your journey involved navigating through options that often included moving east or northeast, with a recent change to north-south movement.

Q:
You are now at (22, 17). The first time you inferred the goal was at (54, -26). The second time you inferred the goal was at (58, -25). The third time you inferred the goal was at (19, -24). Now you infer that the goal is in South 22 steps. According to all your inferences, what are the goal coordinates most likely to be? What is the corresponding goal direction from current position?

A:
"Thought_Q1": "To determine the most likely goal coordinates, consider the reasonable inferences and ignore the outlier. The first inference is (54, -26), the second is (58, -25), and the third is significantly different at (19, -24). Ignoring the third inference, I calculate the average of the first two coordinates: ((54 + 58)/2, (-26 -25)/2) = (56, -25.5), which simplifies to (56, -26) as coordinates are typically integers. Additionally, the current position is (22, 17), and the latest inference suggests a southern shift. Thus, moving 22 steps south from (22, 17) gives new coordinates (22, -5). This inference is also an outlier when compared to the other two reasonable estimations. Therefore, the most probable goal coordinates considering the first two inferences are (56, -26)."
"Answer_Q1": "The goal coordinates are most likely to be (56, -26)."
"Thought_Q2": "To determine the direction of the goal (56, -26) from the current position (22, 17), note the change in x and y coordinates. The x-coordinate changes from 22 to 56, an increase of 34, and the y-coordinate changes from 17 to -26, a decrease of 43. This indicates the goal is to the east and south of the current position, specifically more towards the south. Thus, the goal is southeast of the current position, but more towards the south."
"Answer_Q2": "The goal (56, -26) is in Southeast (more towards south) from current position (22, 17)."

Planning
Q:
You are now at (22, 17). Your current connection includes ['North', 'South']. North is at (22, 18), Visited. South is at (22, 16), Unvisited. The goal coordinates are most likely to be (56, -26). The goal (56, -26) is in Southeast (more towards south) from current position (22, 17). In the past 10 steps, you primarily moved in a northeastern direction, transitioning from (14, 14) to (22, 19), and then shifted to a southern direction, moving from (22, 19) to (22, 17). Your journey involved navigating through options that often included moving east or northeast, with a recent change to north-south movement. The plan is ['1. Move South to head directly towards the goal while exploring unvisited areas.', '2. Continue moving South if the path allows, to make progress directly towards the goal.', '3. Switch to East after making significant southward progress to address the eastward component of the goal location.']. Which step of the plan are you currently implementing? According to all information above, should the plan be updated?. If yes, show the new plan. According to your plan and current connection, choose one in ['North', 'South'] as your next action.

A:
"current_state": "Step 1"
"yes_or_no": "No"
"action_reason": "Currently at an unvisited south connection, which aligns with the plan to move directly towards the goal's more southern location."
"action": "South"
\end{Verbatim}

\subsection{Prompt of plain PReP}
\begin{Verbatim}[fontsize=\tiny, frame=single, samepage=false, breaklines=true, breakanywhere=true]
Q:
You are now at (22, 17). Your current connection includes ['North', 'South']. North is at (22, 18), Visited. South is at (22, 16), Unvisited. Now you infer that the goal is in South 22 steps.
According to all information above, choose one in ['North', 'South'] as your next action.

A:
"action_reason": "Currently at an unvisited south connection, you should explore the unvisited area."
"action": "South"
\end{Verbatim}

\subsection{Prompt of PReP without Reflection}
\begin{Verbatim}[fontsize=\tiny, frame=single, samepage=false, breaklines=true, breakanywhere=true]
Q:
You are now at (22, 17). Your current connection includes ['North', 'South']. North is at (22, 18), Visited. South is at (22, 16), Unvisited. Now you infer that the goal is in South 22 steps. The last plan is ['1. Move South to head directly towards the goal while exploring unvisited areas.', '2. Continue moving South if the path allows, to make progress directly towards the goal.', '3. Switch to East after making significant southward progress to address the eastward component of the goal location.'].
Which step of the plan are you currently implementing? According to all information above, should the plan be updated?. If yes, show the new plan. According to your plan and current connection, choose one in ['North', 'South'] as your next action.

A:
"current_state": "Step 1"
"yes_or_no": "No"
"action_reason": "Currently at an unvisited south connection, which aligns with the plan to move directly towards the goal's more southern location."
"action": "South"
\end{Verbatim}

\subsection{Prompt of PReP without Planning}
\begin{Verbatim}[fontsize=\tiny, frame=single, samepage=false, breaklines=true, breakanywhere=true]
Q:
You are now at (22, 17). Your current connection includes ['North', 'South']. North is at (22, 18), Visited. South is at (22, 16), Unvisited. The goal coordinates are most likely to be (56, -26).The goal (56, -26) is in Southeast (more towards south) from current position (22, 17). In the past 10 steps, you primarily moved in a northeastern direction, transitioning from (14, 14) to (22, 19), and then shifted to a southern direction, moving from (22, 19) to (22, 17). Your journey involved navigating through options that often included moving east or northeast, with a recent change to north-south movement.
According to all information above, choose one in ['North', 'South'] as your next action. 

A:
"action_reason": "Currently at an unvisited south connection, you should explore the unvisited area."
"action": "South"
\end{Verbatim}

\subsection{Prompt of CoT}
\begin{Verbatim}[fontsize=\tiny, frame=single, samepage=false, breaklines=true, breakanywhere=true]
Q:
You are now at (0, 0). Your current connection includes ['Northeast', 'Southwest']. Northeast is at (1, 1), Unvisited. Southwest is at (-1, -1), Unvisited. Now you infer that the goal is in South 51 steps, East 51 steps. According to all information above, choose one in ['Northeast', 'Southwest'] as your next action.

A:
"Thinking": "Starting at (0, 0), the inferred goal is located 51 steps South and 51 steps East, which suggests a general Southeast direction. The available movements are 'Northeast' and 'Southwest'. Moving 'Northeast' to (1, 1) takes us further from the goal directionally, as it increases both North and East coordinates. Conversely, moving 'Southwest' to (-1, -1) also moves away from the goal in terms of latitude but brings us closer longitudinally. Given that both options initially seem counterproductive, the lesser of two deviations would be 'Southwest', because it at least aligns with the southern component of our goal's direction."
"action": "Southwest"
\end{Verbatim}

\subsection{Prompt of InnerMonologue}
\begin{Verbatim}[fontsize=\tiny, frame=single, samepage=false, breaklines=true, breakanywhere=true]
Q:
Scene:
You are now at (0, -27). Your current connection includes ['Southeast', 'Southwest', 'Northeast', 'North']. Southeast is at (1, -28), Unvisited. Southwest is at (-1, -28), Unvisited. Northeast is at (1, -26), Unvisited. North is at (0, -26), Visited. Now you infer that the goal is in South 14 steps.
Planner:
Head towards (0, -48) by moving in the general direction of South when possible.
Accoding to all information above, should the plan be updated?. If yes, show the new plan. According to your plan and current connection, choose one in ['Southeast', 'Southwest', 'Northeast', 'North'] as your next action. 

A:
"Robot Thought": "Continue moving South to reach the goal at (0, -48)."
"Successful Action": "False"
"yes_or_no": "yes"
"new_plan": "Head towards (0, -48) by moving in the general direction of South when possible. Since South is not directly available, choose Southwest as the closest direction to South."
"action": "Southwest"
\end{Verbatim}

\subsection{Prompt of ProgPrompt}
\begin{Verbatim}[fontsize=\tiny, frame=single, samepage=false, breaklines=true, breakanywhere=true]
Q:
You are now at (25, 4). Your current connection includes ['West', 'East']. West is at (24, 4), Visited. East is at (26, 4), Unvisited. Now you infer that the goal is in Southeast 23 steps. The plan is:
def search_goal():
    # 1: Move North because you can't directly move East\n 
    walk('North')
    # 2: If you are at an intersection, you should move East because the goal is in the East
	assert('at intersection')
else: walk('North')
    walk('East')
    # 3: If you are at an intersection again, you should move South because you first move Northwhich deviates
    # from the goal in y-coordinate
    assert('at intersection')
else: walk('East')
    walk('South')
According to all information above, should the plan be updated?. If yes, show the new plan. According to your plan and current connection, choose one in ['West', 'East'] as your next action.

A:
"yes_or_no": "yes"
"new_plan":
"def search_goal():
    # Since the goal is inferred to be at (45, -41) and the current connection includes 'Southwest', 'East'.
    # 'Southwest' has been visited and does not lead closer to the goal direction. 'East' moves closer in the x- 
    # coordinate and is unvisited. Choose 'East' for exploration as it aligns better with the goal's direction and 
    # is unvisited.
walk('East')"
"action": "East"
\end{Verbatim}

\subsection{Prompt of CaP}
\begin{Verbatim}[fontsize=\tiny, frame=single, samepage=false, breaklines=true, breakanywhere=true]
Q:
You are now at (12, 4). Your current connection includes ['Southwest', 'East']. Southwest is at (11, 3), Visited. East is at (13, 4), Unvisited. Now you infer that the goal is in East 22 steps. The plan is:
def search_goal():
    # step 1: Move North until an intersection because you can't directly move East
    while not get_to('intersection'):
walk('North')
    # step 2: If you are at an intersection, you should move East because the goal is in the East	
    if get_to('intersection'):
assert can_go('East'), walk('North)
walk('East')
    # step 3: If you are at an intersection again, you should move South because you first move North which
    # deviates from the goal in y-coordinate
    if get_to('intersection'):
assert can_go('South'), walk('East')
walk('South')
According to all information above, should the plan be updated?. If yes, show the new plan. According to your plan and current connection, choose one in ['Southwest', 'East'] as your next action.

A:
"yes_or_no": "yes"
"new_plan":
"def explore_unknown():
    # step 1: Choose a direction from the current connections that leads to an unvisited location
    if can_go(\'East\'):
walk(\'East\')
    elif can_go(\'Northeast\'):
walk(\'Northeast\')
    elif can_go(\'Southeast\'):
walk(\'Southeast\')
    elif can_go(\'Southwest\'):
walk(\'Southwest\')"
"action": "walk(\'East\')"
\end{Verbatim}

\subsection{Prompt of DEPS}
\begin{Verbatim}[fontsize=\tiny, frame=single, samepage=false, breaklines=true, breakanywhere=true]
Q:
You are now at (18, -8). Your current connection includes ['Southeast', 'Northwest']. Southeast is at (19, -9), Visited. Northwest is at (17, -7), Visited.
Planner:['1. From the current location, move to the nearest unvisited area by choosing a different direction;', '2. Continue moving in this new direction as long as unvisited areas are available and it brings you closer to the goal;', '3. If the goal is not found and no further paths in the new direction are available, reassess the position and plan further steps based on the remaining unvisited areas.']
According to all information above, should the plan be updated? If yes, show the new plan. According to your plan and current connection, choose one in ['Southeast', 'Northwest'] as your next action.

A:
"Descriptor": "I failed on step 1 'From the current location, move to the nearest unvisited area by choosing a different direction'. I'm now at (18, -8)."  "Explainer": "It's because both available directions, Southeast and Northwest, lead to previously visited locations. I need to choose a new direction to find unvisited areas."  
"yes_or_no": "yes"
"new_plan": ["1. From the current location, move to the nearest unvisited area by choosing a direction not previously available; 2. Continue moving in this new direction as long as unvisited areas are available and it brings you closer to the goal; 3. If the goal is not found and no further paths in the new direction are available, reassess the position and plan further steps based on the remaining unvisited areas."]
"action": "Northwest"
\end{Verbatim}

\end{document}